\documentclass[11pt,a4paper]{article}
\usepackage[hyperref]{emnlp2018}
\usepackage{times}
\usepackage{latexsym}

\usepackage{url}
\usepackage{graphicx}
\usepackage{CJKutf8}
\usepackage{siunitx}
\usepackage{amsmath}

\aclfinalcopy % Uncomment this line for the final submission

\title{Live Video Comment Generation Based on Surrounding Frames and Live Comments}

\author{Damai Dai \\
  School of EECS, Peking University \\
  {\tt daidamai@pku.edu.cn}
}

\date{\today}

\begin{document}
\maketitle
\begin{abstract}
In this paper, we propose the task of live comment generation. Live comments are a new form of comments on videos, which can be regarded as a mixture of comments and chats. A high-quality live comment should be not only relevant to the video, but also interactive with other users. In this work, we first construct a new dataset for live comment generation. Then, we propose a novel end-to-end model to generate the human-like live comments by referring to the video and the other users' comments.
% To train the model, We collect a large amount of live comments and their corresponding frames. 
Finally, we evaluate our model on the constructed dataset. Experimental results show that our method can significantly outperform the baselines.\footnote{The dataset and the code will be released to the public if the manuscript is accepted.} 
\end{abstract}

\setlength{\abovedisplayskip}{3pt}
\setlength{\belowdisplayskip}{3pt}
\setlength{\abovedisplayshortskip}{3pt}
\setlength{\belowdisplayshortskip}{3pt}

\section{Introduction}

\begin{figure}[t]
\centering
\includegraphics[scale=0.2]{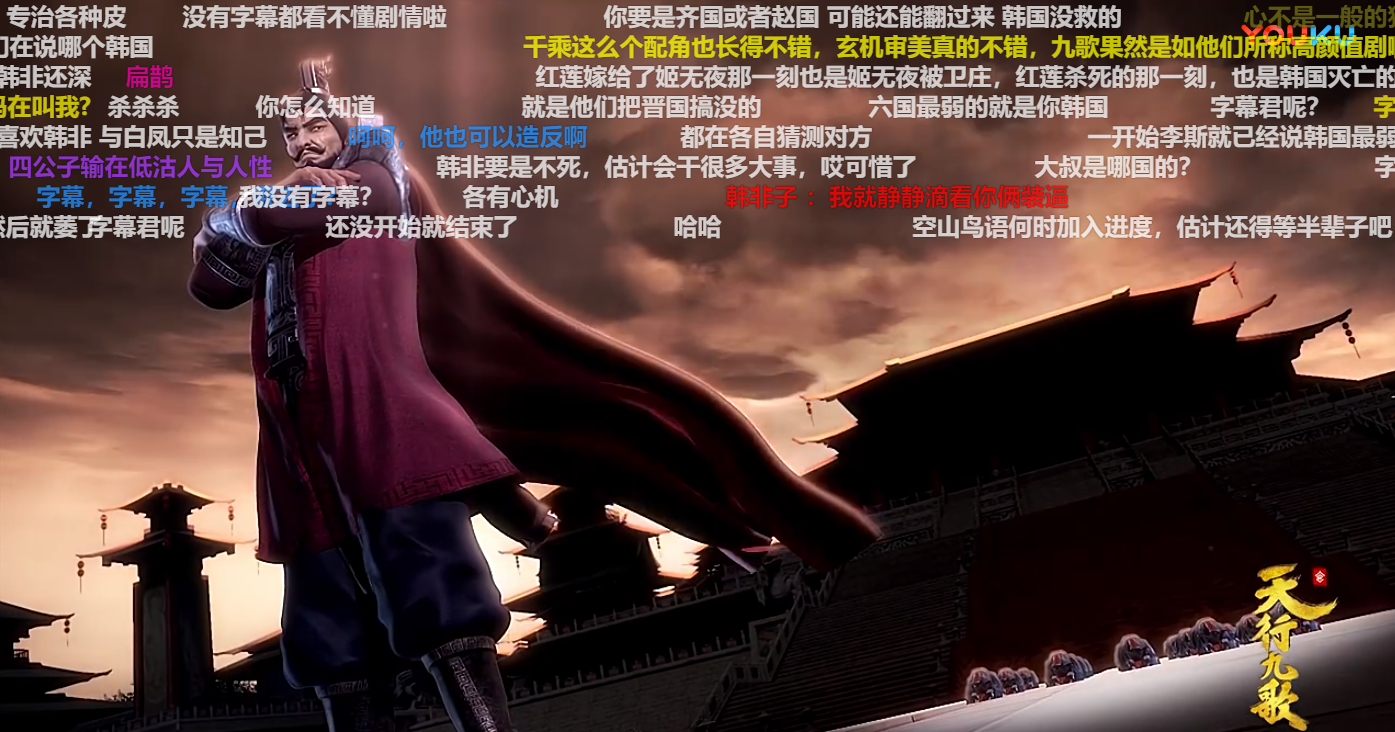}
\caption{An example of live comments. The colorful Chinese characters are the real-time live comments.}
\label{fig:live commenting}
\end{figure}

In this paper, we focus on the task of automatically generating live comments. Live comments (also known as ``\begin{CJK}{UTF8}{gbsn}弹幕 \end{CJK}bullet screen'' in Chinese) are a new form of comments that appear in videos. We show an example of live comments in Figure~\ref{fig:live commenting}. Live comments are popular among youngsters as it plays the role of not only sharing opinions but also chatting. Automatically generating live comments can make the video more interesting and appealing. 

%Different from the traditional comments on videos, a live comment appears at a certain point of the video timeline, which gives it some unique characteristics. First, the live comments can be causal chats instead of serious comments on the content of the video. For example, in Figure~\ref{fig:live commenting}, one of the live comments is ``\begin{CJK}{UTF8}{gbsn}我没有字幕?\end{CJK} (No comments?)''. Second, many live comments are related to the contexts around an exact moment instead of the whole video. 

Different from the other video-to-text tasks, such as video caption, a live comment appears at a certain point of the video timeline, which gives it some unique characteristics. The live comments can be causal chats about a topic with other users instead of serious descriptions of the videos. Therefore, a human-like comment should be not only relevant to the video, but also interactive with other users.

\begin{figure*}[t]
\centering
\includegraphics[scale=0.5]{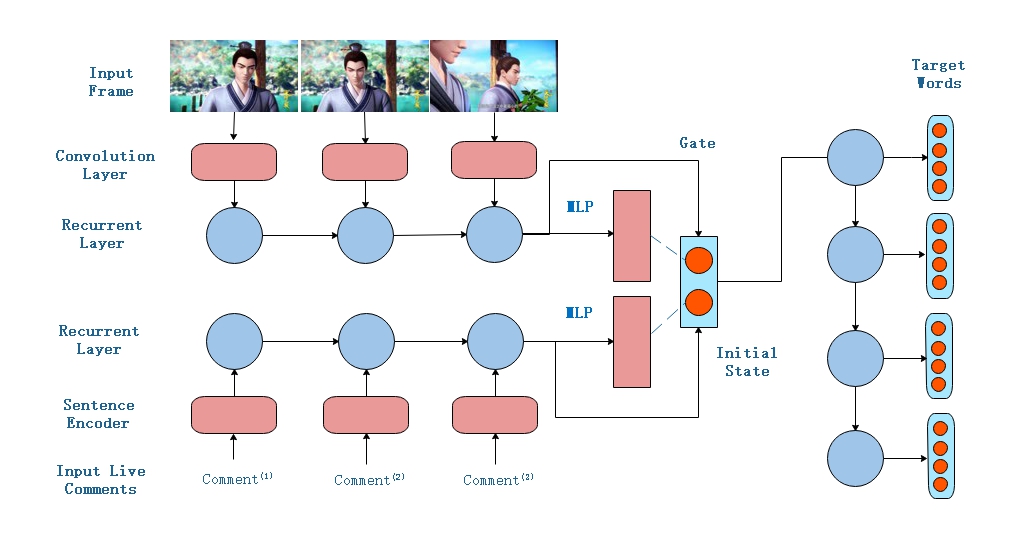}
\caption{An illustration of our joint video and live comment model. We make use of not only the surrounding frames but also the surrounding live comments to generate the target live comment. 
% We use CNN to encode each frame and RNN to encode each live comment.
}
\label{fig:model}
\vspace{-0.1in}
\end{figure*}

In this paper, we aim at generating human-like live comments for the videos. We propose a novel end-to-end model to generate the comments by referring to the video and other users' comments. %In our model, we use a CNN to extract information from the video frames as well as a RNN to encode the live comments. 
%In this paper, we also apply the encoder-decoder framework to deal with the problem of live comment generation. We propose a \textbf{joint video and live comment encoder}, where CNN is used to extract information from the video frames and RNN is used to encode the live comments. 
We have access to not only the current frame but also the surrounding frames and live comments because a live comment and its associated frame are in the context of a series of surrounding frames and live comments. To make use of the information in those two parts, we design a model that encodes the surrounding frames and live comments together into a vector, based on which we decode the new live comment. Experimental results show that our model can generate human-like live comments.
% Because of the difference of task setting, a live comment and its associated frame are not isolated, but in the context of a series of surrounding frames and live comments. Therefore, we have access to not only the current frame but also the surrounding frames and live comments. To make use of the information in those two parts, we design a model that encodes the surrounding frames and live comments together into a vector, based on which we decode the new live comment. Experimental results show that our model can generate human-like live comments.

%Generating descriptions based on the scenes in a video has also been studied \citep{gerber1996knowledge,yao2010i2t:}. However, these works all generate sentences based on rules and the scenario they deal with tends to be simple. 

Our contributions are two folds:
 \begin{itemize}
\item We propose a new task of automatically generating live comments and %crawl live comments and the corresponding video frames from the Internet to 
build a dataset with videos and live comments for live comment generation.
\item We propose a novel joint video and live comment model to make use of the current frame, the surrounding frames, and the surrounding live comments to generate a new live comment. Experimental results show that our model can generate human-like live comments.
\end{itemize}

\section{Proposed Model}\label{model}

In Figure~\ref{fig:model} we show the architecture of our model. Our live comment generation model is composed of four parts: a video encoder, a text encoder, a gated component, and a live comment generator. The video encoder encodes $n$ consecutive frames, and the text encoder encodes $m$ surrounding live comments into the vectors. The gated component aggregates the video and the comments into a joint representation. Finally, the live comment generator generates the target live comment.

\subsection{Video Encoder}

In our task, each generated live comment is attached with $n$ consecutive frames. In the video encoding part, each frame $f^{(i)}$ is first encoded into a vector $v_{f}^{(i)}$ by a convolution layer. We then use a GRU~\cite{cho2014learning} layer to encode all the frame vectors into their hidden states $h_{v}^{(i)}$:
\begin{align}
v_{f}^{(i)} &= CNN(f^{(i)}) \\
h_{v}^{(i)} =& GRU(v_{f}^{(i)}, h_{v}^{(i-1)})
\end{align}
We set the last hidden state $h_{v}^{(n)}$ as the representation of the video $v_v = h_v^{(n)}$.

\subsection{Text Encoder}

In the comment encoding part, a live comment $c^{(i)}$ with $L^{(i)}$ words $(w_{1}^{(i)},w_{2}^{(i)},\cdots,w_{L^{(i)}}^{(i)})$ is first encoded into a series of word-level hidden states ($h_{w_{1}}^{(i)},h_{w_{2}}^{(i)},\cdots,h_{w_{L^{(i)}}}^{(i)}$),  using a word-level GRU layer. We use the last hidden state $h_{L^{(i)}}^{(i)}$ as the representation for $c^{(i)}$  denoted as $v_{c}^{(i)}$. Then we use a sentence-level GRU layer to encode all the live comment vectors of different live comments into their hidden states $h_{c}^{(i)}$:
\begin{eqnarray}
h_{w_{j}}^{(i)} = GRU(w_{j}^{(i)}, h_{w_{j-1}}^{(i)})\\
%v_{c}^{(i)} = h_{L^{(i)}}^{(i)}\\
h_c^{(i)} = GRU(v_{c}^{(i)}, h_c^{(i-1)})
%v_c = h_c^{(m)}\\
\end{eqnarray}
The last hidden state $h_{c}^{(m)}$ is used as the vector of all the live comments $v_c = h_c^{(m)}$.

\subsection{Gated Selection}

In order to describe how much information we should get from the video and the live comments, we apply a gated multi-layer perceptron (MLP) to combine $v_{c}$ and $v_{v}$, and get the final vector $h$:
%We use ReLU\cite{nair2010rectified} as the non-linearity function. 
%A $softmax$ layer is used to determine the percentage of the two parts.
\begin{eqnarray}
s_{v} = u\mathop{ReLU}(W_{v}v_{v}+b_{v}) \\
% \end{equation}
% \begin{equation}
s_{c} = u\mathop{ReLU}(W_{c}v_{c}+b_{c}) \\
h = 
\left[ 
	\begin{array}{cc}
  		\frac{e^{s_c}}{e^{s_v}+e^{s_c}}v_c, & \frac{e^{s_v}}{e^{s_v}+e^{s_c}}v_v \\
  	\end{array} 
\right ]
\end{eqnarray}
where $u,W,b$ are trainable parameters.

\subsection{Live Comment Generator}

We use a GRU to decode the live comment. The encoder encodes the frames and live comments jointly into a vector $h$. The probability of generating a sentence given the encoded vector $h$ is defined as,
\begin{equation}
p(w_0,...,w_T|h) = \prod_{t=1}^{T}p(w_t|w_0,...,w_{t-1},h)
\end{equation}
More specifically, the probability distribution of word $w_{i}$ is calculated as follows,
\begin{align}
h_i =& GRU(w_{i}, h_{i-1})\\
p(w_{i}|w_0,...,w_{i-1},&h) = softmax(W_{o}h_i)
\end{align}

%\subsection{Training and Loss Function}

%The loss function consists of two parts, which are the cross entropy loss of summarization and that of sentiment classification:
%\begin{equation}
%L_{s}=-\sum_t{{y_t}\log{p(y_t|x)}}
%\end{equation}
%\begin{equation}
%L_{c}=-{l}\log{p(l|x)}
%\end{equation}
%where ${y_t}$ and ${l}$ are the ground truth of words and labels, and $p(y_t|x)$ and $p(l|x)$ are the probability distribution of words and labels. 

\section{Experiments}

In this section, we show the experimental results of our proposed model and compare it with three baselines on the dataset we construct from Youku.\footnote{\url{http://www.youku.com}}

\subsection{Live Comment Dataset Construction}
% use ffmpeg\footnote{see \url{http://ffmpeg.org/} for details} to
\noindent\textbf{Video frames: }We extract frames from an animated TV series named ``Tianxingjiuge'' \begin{CJK}{UTF8}{gbsn} (``天行九歌'')\end{CJK} at a frequency of 1 frame per second. We get \num{21600} frames from 40 videos in total, with a shape of $128 \times 72$ for each frame. We split the frames into training set (\num{21000}) and test set (600). 

\noindent\textbf{Live comments: } We use the developer tools in Google Chrome to manually detect the live comments sources, via which we get all live comments of the 40 videos. For each extracted frame, we select 5 live comments which are the nearest to the frame at the time they appear. 

\noindent\textbf{Reference set: } Besides the live comments in our training set and test set, we crawl \num{1036978} extra live comments to be the reference set for calculating BLEU score and perplexity which can evaluate the fluency of generated live comments (refer to Table~\ref{bleu_ppl}). 

\noindent\textbf{Copyright statement: } The dataset we construct can only be used for scientific research. The copyright belongs to the original website \textbf{Youku}.

\subsection{Baselines}

Besides the model described in section ~\ref{model}, we have three baseline methods: 

\noindent\textbf{Frame-to-Comment (F2C)}~\citep{vinyals2015show} applies a CNN to encode the current frame to a vector, based on which the decoder generates the target live comment. 

\noindent\textbf{Moment-to-Comment (M2C)} applies an RNN to make use of one live comment near the current frame besides the CNN for the frame. The two encoded vectors are concatenated to be the initial hidden state for the decoder.

\noindent\textbf{Context-to-Comment (C2C)} is similar to ~\citep{venugopalan2015sequence} which makes use of a series of surrounding frames and live comments by encoding them with extra higher-level RNNs. 
% This model is similar to our proposed model except that the gated mechanism is not applied. 

\subsection{Evaluation Metrics}

We design two types of evaluation metrics: \textbf{human evaluation} and \textbf{automatic evaluation}. 

\noindent\textbf{Human Evaluation: }We evaluate in three aspects:
% \item \textbf{Novelty} is designed to measure the interestingness of the generated live comments
\textbf{Fluency} is designed to measure whether the generated live comments are fluent setting aside the relevance to videos. 
\textbf{Relevance} is designed to measure the relevance between the generated live comments and their associated frames. 
\textbf{Overall Score} is designed to synthetically measure the confidence that the generated live comments are made by humans in the context of the video. 
For all of the above three aspects, we stipulate the score to be an integer in $\{1, 2, 3, 4, 5\}$. The higher the better. The scores are evaluated by three seasoned native speakers and finally we take the average of three raters as the final result. 

\noindent\textbf{Automatic Evaluation: }We adopt two metrics: \textbf{BLEU score} \cite{Papineni:2002:BMA:1073083.1073135} and \textbf{perplexity}. These two metrics are designed to measure whether the generated live comments accord with the human-like style. 
To get \textbf{BLEU scores}, for each generated live comment, we calculate its BLEU-4 score with all live comments in the reference set, and then we pick the maximal one to be its final score.
\textbf{Perplexity} is to measure the language quality of the generated live comments, which is estimated as, $$perplexity = 2^{-\frac{1}{n}\sum_{i}\log{p(w_i \vert h_i)}}$$ for each word $w_i$ in the sentence, $h_i$ is the predicted word.

\subsection{Experimental Details}

The vocabulary is limited to be the 34,100 most common words in the training dataset. We use a shared embedding between encoder and decoder and set the word embedding size to 300. The word embedding is randomly initialized and learned automatically during the training.

For the 3 GRUs used in the encoding stage, we set the hidden size to 300. For the decoding GRU, we set the hidden size to 600. For the encoding CNN, we use 3 convolution layers and 3 linear layers, and get a final vector with a size of 300. The batch size is 512.
% with kernel sizes of 9, 5, 5 respectively. After the first two convolution layers, we use a max-pooling layer with a kernel size of 2. After the third convolution layer, we use a MLP to get a final vector with a size of 300. 
%
% For the gate MLP, we use two different and independent MLPs to process image hidden vector and live comment hidden vector, and finally get two scalars as their weights. 
%
We use the Adam~\cite{KingmaBa2014} optimization method to train the model. 
For the hyper-parameters of Adam optimizer, we set the learning rate $\alpha = 0.0003$, two momentum parameters $\beta_{1} = 0.9$ and $\beta_{2} = 0.999$
respectively, and $\epsilon = 1 \times 10^{-8}$. 
During training, we use ``teacher forcing''~\cite{Williams1989Experimental} to make our model converge faster and we set the teacher forcing ratio $p = 0.5$. 

% and we train the model for 20000 iterations. 
\subsection{Results and Analysis}

\begin{table}[tb]
	\centering
	\normalsize
    \setlength{\tabcolsep}{2.0mm}{
        \begin{tabular}{l|c|c|c}
            \hline
            \textbf{Model} & \textbf{Fluency} & \textbf{Relevance} & \textbf{Overall} \\ \hline
            F2C  & 3.80 & 1.73 & 2.56 \\
            M2C & 3.88 & 1.74 &  2.58 \\
            C2C & 4.11 & 1.83 &  2.94 \\
            \textbf{Proposal}  & \textbf{4.45} & \textbf{1.95} & \textbf{3.30} \\ \hline
            Human & 4.84 & 2.60 & 3.87 \\ \hline
        \end{tabular}
    }
	\caption{Results of human evaluation metrics on the test set. \textit{Human} means the real-world live comments from videos. \textit{Overall} is a comprehensive metric given by our annotators.%The Spearman's correlation coefficients between any two raters are all near 0.6 and at an average of 0.63. 
    }\label{manual}
    \vspace{-0.1in}
\end{table}

% \begin{table}[tb]
% 	\centering
% 	\normalsize
%     \setlength{\tabcolsep}{1.0mm}{
%         \begin{tabular}{@{}l@{}|@{}c@{}|@{}c@{}|@{}c@{}|@{}c@{}}
%             \hline
%             \textbf{Model} & \textbf{Fluency} & \textbf{Relevance} & \textbf{Probability} & \textbf{All} \\ \hline
%             F2C  & 3.80 & 1.73 & 2.56 & 2.70 \\
%             M2C & 3.88 & 1.74 &  2.58 & 2.73 \\
%             C2C & 4.11 & 1.83 &  2.94 & 2.96 \\
%             \textbf{Proposal}  & \textbf{4.45} & \textbf{1.95} & \textbf{3.30} & \textbf{3.23} \\ \hline
%             Human & 4.84 & 2.60 & 3.87 & 3.77
%         \end{tabular}
%     }
% 	\caption{Results of manual evaluation metrics on the test set. The ``all'' score is the average of the other three scores. ``Human'' means the evaluation result on real live comments. The Spearman's correlation coefficients between any two raters are all near 0.6, and are at an average of 0.63. }\label{manual}
% \end{table}

As shown in Table~\ref{manual}, our model achieves the highest scores over the baseline models in all three degrees. When only the current frame or one extra live comment is considered (\textbf{F2C} and \textbf{M2C}), the generated live comments have low scores. After considering more surrounding frames and live comments (\textbf{C2C}), all of the scores get higher. Finally, with the gate mechanism that can automatically decide the weights of surrounding frames and live comments, our \textbf{proposal} achieves the highest scores, which are almost near to those of real-world live comments. We use Spearman's Rank correlation coefficients to evaluate the agreement among the raters. The coefficients between any two raters are all near 0.6 and at an average of 0.63. These high coefficients show that our human evaluation scores are consistent and credible. 
% This observation also accords with the example we show in Table~\ref{exmpl}. The live comment generated by our \textbf{final model} can vividly reflect the ambitious character of the person who wants to conquer other countries in the associated frame shown in Figure~\ref{fig:live commenting}. 

\noindent\textbf{Relevance: }The relevant scores presented in Table~\ref{manual} show that the live comments generated by all models do not achieve high relevant scores, which means that many of the generated live comments are not relevant to the current frame. We go through the real live comments and find that about 75.7\% live comments are not relevant to the current frame, but are just chatting. In fact, we can find from Table~\ref{manual} that the relevance score of real live comments is not high as well. Therefore, the low relevant scores are reasonable. Still, our proposal can generate more relevant live comments owing to its ability to combine the information from the surrounding frames and live comments.

\noindent\textbf{Fluency: }From the fluency score presented in Table~\ref{manual}, the BLEU-4 score and the perplexity score presented in Table \ref{bleu_ppl}, we can see our proposal can generate live comments which best accord with the human-like style. 
% with the highest degree of fluency. 

\noindent\textbf{Informativeness: }From the \textit{Average Length}  in Table~\ref{bleu_ppl}, we can see our proposal improves the length of the generated live comments, which indicates that more meaningful information is embodied. 

% The progressive improvement in the \textbf{total score} shown in Table~\ref{manual} shows that our final model performs the best, and we can intuitively recognize this from the example in Table~\ref{exmpl} . 

\begin{table}[tb]
	\centering
	\normalsize
    \setlength{\tabcolsep}{0.5mm}{
	\begin{tabular}{l|c|c|c}
		\hline
		\textbf{Model} & \textbf{BLEU-4} & \textbf{Perplexity} & \textbf{Average Length} \\ \hline
		F2C & 63.8 & 181.15 & 4.50 \\	
		M2C & 65.4 & 146.15 & 4.28 \\
		C2C & 77.2 & 93.82 & 4.72 \\
		\textbf{Proposal}  & \textbf{83.5} & \textbf{56.85} & \textbf{5.17} \\ \hline
    	Human & 97.9 & 41.01 & 5.86 \\ \hline
    \end{tabular}
    }
	\caption{Automatic evaluation results of the BLEU-4 scores, perplexities and average lengths of live comments on the test set. }\label{bleu_ppl}
    \vspace{-0.1in}
\end{table}

\section{Related Work}

Inspired by the great success achieved by the sequence-to-sequence learning framework in machine translation \citep{sutskever2014sequence,cho2014learning,bahdanau2014neural}, \citet{vinyals2015show} and \citet{mao2014explain} proposed to use a deep convolutional neural network (CNN) to encode the image and a recurrent neural network (RNN) to generate the image captions. \citet{xu2015show} further proposed to apply attention mechanism to focus on certain parts of the image when decoding. Using CNN to encode the image while using RNN to decode the description is natural and effective when generating textual descriptions. 

One task that is similar to live comment generation is image caption generation, which is an area that has been studied for a long time. \citet{farhadi2010every} tried to generate descriptions of an image by retrieving from a big sentence pool. \citet{kulkarni2011baby} proposed to generate descriptions based on the parsing result of the image with a simple language model. These systems are often applied in a pipeline fashion, and the generated description is not creative. More recent work is to use stepwise merging network to improve the performance \cite{liu2018}.

Another task which is similar to this work is video caption generation. \citet{venugopalan2015sequence} proposed to use CNN to extract image features, and use LSTM to encode them and decode a sentence. Similar models \citep{shetty2016frame,jin2016describing,ramanishka2016multimodal,dong2016early,pasunuru2017multi,shen2017weakly} are also proposed to handle the task of video caption generation. \citet{Das2017ICCV,Das2017CVPR} introduce the task of Visual Dialog, which requires an AI agent to answer a question about an image when given the image and a dialog history. 

We cast this problem as a natural language generation problem, and we are also inspired by the recent related work of natural language generation models with the text inputs \cite{DBLP:journals/corr/abs-1803-01465,DBLP:journals/corr/abs-1805-04871,DBLP:journals/corr/abs-1805-05181,DBLP:journals/corr/abs-1802-01345}. 

\section{Conclusion}

In this paper, we propose the task of live comment generation. In order to generate high-quality comments, we propose a novel neural model which makes use of the surrounding frames in the video and other surrounding live comments. Experimental results show that our model performs better than the baselines in various metrics, and even approaches the performance of human.

% include your own bib file like this:
\bibliographystyle{acl_natbib_nourl}
\bibliography{emnlp2018}

\begin{thebibliography}{25}
\expandafter\ifx\csname natexlab\endcsname\relax\def\natexlab#1{#1}\fi

\bibitem[{Bahdanau et~al.(2014)Bahdanau, Cho, and Bengio}]{bahdanau2014neural}
Dzmitry Bahdanau, Kyunghyun Cho, and Yoshua Bengio. 2014.
\newblock Neural machine translation by jointly learning to align and
  translate.
\newblock \emph{arXiv preprint arXiv:1409.0473}.

\bibitem[{Cho et~al.(2014)Cho, Van~Merri{\"e}nboer, Gulcehre, Bahdanau,
  Bougares, Schwenk, and Bengio}]{cho2014learning}
Kyunghyun Cho, Bart Van~Merri{\"e}nboer, Caglar Gulcehre, Dzmitry Bahdanau,
  Fethi Bougares, Holger Schwenk, and Yoshua Bengio. 2014.
\newblock Learning phrase representations using rnn encoder-decoder for
  statistical machine translation.
\newblock \emph{arXiv preprint arXiv:1406.1078}.

\bibitem[{Das et~al.(2017{\natexlab{a}})Das, Kottur, Gupta, Singh, Yadav,
  Moura, Parikh, and Batra}]{Das2017CVPR}
Abhishek Das, Satwik Kottur, Khushi Gupta, Avi Singh, Deshraj Yadav,
  Jos{\'{e}}~M.F. Moura, Devi Parikh, and Dhruv Batra. 2017{\natexlab{a}}.
\newblock {Visual dialog}.
\newblock In \emph{Proceedings - 30th IEEE Conference on Computer Vision and
  Pattern Recognition, CVPR 2017}, volume 2017-January, pages 1080--1089.

\bibitem[{Das et~al.(2017{\natexlab{b}})Das, Kottur, Moura, Lee, and
  Batra}]{Das2017ICCV}
Abhishek Das, Satwik Kottur, Jose~M.F. Moura, Stefan Lee, and Dhruv Batra.
  2017{\natexlab{b}}.
\newblock {Learning Cooperative Visual Dialog Agents with Deep Reinforcement
  Learning}.
\newblock In \emph{Proceedings of the IEEE International Conference on Computer
  Vision}, volume 2017-October, pages 2970--2979.

\bibitem[{Dong et~al.(2016)Dong, Li, Lan, Huo, and Snoek}]{dong2016early}
Jianfeng Dong, Xirong Li, Weiyu Lan, Yujia Huo, and Cees~GM Snoek. 2016.
\newblock Early embedding and late reranking for video captioning.
\newblock In \emph{Proceedings of the 2016 ACM on Multimedia Conference}, pages
  1082--1086. ACM.

\bibitem[{Farhadi et~al.(2010)Farhadi, Hejrati, Sadeghi, Young, Rashtchian,
  Hockenmaier, and Forsyth}]{farhadi2010every}
Ali Farhadi, Mohsen Hejrati, Mohammad~Amin Sadeghi, Peter Young, Cyrus
  Rashtchian, Julia Hockenmaier, and David~A Forsyth. 2010.
\newblock Every picture tells a story: generating sentences from images.
\newblock pages 15--29.

\bibitem[{Jin et~al.(2016)Jin, Chen, Chen, Xiong, and
  Hauptmann}]{jin2016describing}
Qin Jin, Jia Chen, Shizhe Chen, Yifan Xiong, and Alexander Hauptmann. 2016.
\newblock Describing videos using multi-modal fusion.
\newblock In \emph{Proceedings of the 2016 ACM on Multimedia Conference}, pages
  1087--1091. ACM.

\bibitem[{Kingma and Ba(2014)}]{KingmaBa2014}
Diederik~P. Kingma and Jimmy Ba. 2014.
\newblock Adam: {A} method for stochastic optimization.
\newblock \emph{CoRR}, abs/1412.6980.

\bibitem[{Kulkarni et~al.(2011)Kulkarni, Premraj, Dhar, Li, Choi, Berg, and
  Berg}]{kulkarni2011baby}
Girish Kulkarni, Visruth Premraj, Sagnik Dhar, Siming Li, Yejin Choi,
  Alexander~C Berg, and Tamara~L Berg. 2011.
\newblock Baby talk: Understanding and generating simple image descriptions.
\newblock pages 1601--1608.

\bibitem[{Liu et~al.(2018)Liu, Ren, Liu, Wang, and Sun}]{liu2018}
Fenglin Liu, Xuancheng Ren, Yuanxin Liu, Houfeng Wang, and Xu~Sun. 2018.
\newblock Stepwise image-topic merging network for generating detailed and
  comprehensive image captions.
\newblock In \emph{EMNLP 2018}.

\bibitem[{Ma et~al.(2018{\natexlab{a}})Ma, Sun, Li, Li, Li, and
  Ren}]{DBLP:journals/corr/abs-1803-01465}
Shuming Ma, Xu~Sun, Wei Li, Sujian Li, Wenjie Li, and Xuancheng Ren.
  2018{\natexlab{a}}.
\newblock Query and output: Generating words by querying distributed word
  representations for paraphrase generation.
\newblock \emph{CoRR}, abs/1803.01465.

\bibitem[{Ma et~al.(2018{\natexlab{b}})Ma, Sun, Wang, and
  Lin}]{DBLP:journals/corr/abs-1805-04871}
Shuming Ma, Xu~Sun, Yizhong Wang, and Junyang Lin. 2018{\natexlab{b}}.
\newblock Bag-of-words as target for neural machine translation.
\newblock \emph{CoRR}, abs/1805.04871.

\bibitem[{Mao et~al.(2014)Mao, Xu, Yang, Wang, and Yuille}]{mao2014explain}
Junhua Mao, Wei Xu, Yi~Yang, Jiang Wang, and Alan~L Yuille. 2014.
\newblock Explain images with multimodal recurrent neural networks.
\newblock \emph{arXiv: Computer Vision and Pattern Recognition}.

\bibitem[{Papineni et~al.(2002)Papineni, Roukos, Ward, and
  Zhu}]{Papineni:2002:BMA:1073083.1073135}
Kishore Papineni, Salim Roukos, Todd Ward, and Wei-Jing Zhu. 2002.
\newblock Bleu: A method for automatic evaluation of machine translation.
\newblock In \emph{Proceedings of the 40th Annual Meeting on Association for
  Computational Linguistics}, ACL '02, pages 311--318, Stroudsburg, PA, USA.
  Association for Computational Linguistics.

\bibitem[{Pasunuru and Bansal(2017)}]{pasunuru2017multi}
Ramakanth Pasunuru and Mohit Bansal. 2017.
\newblock Multi-task video captioning with video and entailment generation.
\newblock \emph{arXiv preprint arXiv:1704.07489}.

\bibitem[{Ramanishka et~al.(2016)Ramanishka, Das, Park, Venugopalan, Hendricks,
  Rohrbach, and Saenko}]{ramanishka2016multimodal}
Vasili Ramanishka, Abir Das, Dong~Huk Park, Subhashini Venugopalan, Lisa~Anne
  Hendricks, Marcus Rohrbach, and Kate Saenko. 2016.
\newblock Multimodal video description.
\newblock In \emph{Proceedings of the 2016 ACM on Multimedia Conference}, pages
  1092--1096. ACM.

\bibitem[{Shen et~al.(2017)Shen, Li, Su, Li, Chen, Jiang, and
  Xue}]{shen2017weakly}
Zhiqiang Shen, Jianguo Li, Zhou Su, Minjun Li, Yurong Chen, Yu-Gang Jiang, and
  Xiangyang Xue. 2017.
\newblock Weakly supervised dense video captioning.
\newblock In \emph{Proceedings of the IEEE Conference on Computer Vision and
  Pattern Recognition}, volume~2, page~10.

\bibitem[{Shetty and Laaksonen(2016)}]{shetty2016frame}
Rakshith Shetty and Jorma Laaksonen. 2016.
\newblock Frame-and segment-level features and candidate pool evaluation for
  video caption generation.
\newblock In \emph{Proceedings of the 2016 ACM on Multimedia Conference}, pages
  1073--1076. ACM.

\bibitem[{Sutskever et~al.(2014)Sutskever, Vinyals, and
  Le}]{sutskever2014sequence}
Ilya Sutskever, Oriol Vinyals, and Quoc~V Le. 2014.
\newblock Sequence to sequence learning with neural networks.
\newblock In \emph{Advances in neural information processing systems}, pages
  3104--3112.

\bibitem[{Venugopalan et~al.(2015)Venugopalan, Rohrbach, Donahue, Mooney,
  Darrell, and Saenko}]{venugopalan2015sequence}
Subhashini Venugopalan, Marcus Rohrbach, Jeffrey Donahue, Raymond Mooney,
  Trevor Darrell, and Kate Saenko. 2015.
\newblock Sequence to sequence-video to text.
\newblock In \emph{Proceedings of the IEEE international conference on computer
  vision}, pages 4534--4542.

\bibitem[{Vinyals et~al.(2015)Vinyals, Toshev, Bengio, and
  Erhan}]{vinyals2015show}
Oriol Vinyals, Alexander Toshev, Samy Bengio, and Dumitru Erhan. 2015.
\newblock Show and tell: A neural image caption generator.
\newblock In \emph{Proceedings of the IEEE conference on computer vision and
  pattern recognition}, pages 3156--3164.

\bibitem[{Williams and Zipser(1989)}]{Williams1989Experimental}
Ronald~J. Williams and David Zipser. 1989.
\newblock Experimental analysis of the real-time recurrent learning algorithm.
\newblock \emph{Connection Science}, 1(1):87--111.

\bibitem[{Xu et~al.(2018{\natexlab{a}})Xu, Sun, Ren, Lin, Wei, and
  Li}]{DBLP:journals/corr/abs-1802-01345}
Jingjing Xu, Xu~Sun, Xuancheng Ren, Junyang Lin, Bingzhen Wei, and Wei Li.
  2018{\natexlab{a}}.
\newblock {DP-GAN:} diversity-promoting generative adversarial network for
  generating informative and diversified text.
\newblock \emph{CoRR}, abs/1802.01345.

\bibitem[{Xu et~al.(2018{\natexlab{b}})Xu, Sun, Zeng, Ren, Zhang, Wang, and
  Li}]{DBLP:journals/corr/abs-1805-05181}
Jingjing Xu, Xu~Sun, Qi~Zeng, Xuancheng Ren, Xiaodong Zhang, Houfeng Wang, and
  Wenjie Li. 2018{\natexlab{b}}.
\newblock Unpaired sentiment-to-sentiment translation: {A} cycled reinforcement
  learning approach.
\newblock \emph{CoRR}, abs/1805.05181.

\bibitem[{Xu et~al.(2015)Xu, Ba, Kiros, Cho, Courville, Salakhudinov, Zemel,
  and Bengio}]{xu2015show}
Kelvin Xu, Jimmy Ba, Ryan Kiros, Kyunghyun Cho, Aaron Courville, Ruslan
  Salakhudinov, Rich Zemel, and Yoshua Bengio. 2015.
\newblock Show, attend and tell: Neural image caption generation with visual
  attention.
\newblock In \emph{International Conference on Machine Learning}, pages
  2048--2057.

\end{thebibliography}

\end{document}